\def\year{2020}\relax
\documentclass[letterpaper]{article} 
\usepackage{aaai20}  
\usepackage[frozencache,cachedir=.]{minted}
\usepackage{times}  
\usepackage{helvet} 
\usepackage{courier}  
\usepackage[hyphens]{url}  
\usepackage{graphicx} 
\usepackage{graphicx}  
\usepackage{algorithm}
\usepackage{algorithmicx}
\usepackage{algpseudocode}
\usepackage{amsmath,amsfonts,amssymb,bm}
\usepackage{natbib}
\usepackage{pifont} 
\usepackage{booktabs}
\usepackage[colorlinks,urlcolor=blue]{hyperref}

\newcommand{\cmark}{\ding{51}}
\title{Learning a Deep Generative Model like a Program: the Free Category Prior}
\author{Eli Sennesh\textsuperscript{\rm 1} \\ 
\textsuperscript{\rm 1}Khoury College of Computer Science\\ Northeastern University \\ 
440 Huntington Ave\\
Boston, MA 02115\\
sennesh.e@northeastern.edu 
}
\begin{document}

\maketitle

\noindent Humans surpass the cognitive abilities of most other animals in our ability to ``chunk'' concepts into words, and then combine the words to combine the concepts.  In this process, we make ``infinite use of finite means''\footnote{Attributed to Wilhelm von Humboldt}, enabling us to learn new concepts quickly ~\citep{Lake2013} and nest concepts within each-other ~\citep{Lake2019}.  We construct concepts to model the world around us ~\citep{Goodman2014}.  In contrast, in artificial intelligence we still mostly focus on learning discriminative functions mapping from sensory inputs to decision outputs, without modeling the data-generating process ~\citep{Zhu2019}.

While program induction and synthesis remain at the heart of foundational theories of artificial intelligence ~\citep{SOLOMONOFF19641,Vitanyi2000}, only recently has the community moved forward in attempting to use program learning as a benchmark task itself ~\citep{Chollet2019}.  Models of program learning are most often based upon probabilistic context-free grammars ~\citep{Overlan2017,Romano2018}, with neural networks and/or reinforcement learning sometimes used to aid inference ~\citep{Ellis2018,ellis2020dreamcoder}.  However, not every grammatical program in the support of a PCFG yields a meaningful likelihood.  Indeed, a recent experiment by \citet{Bramley2018} found that 35\% of sampled hypotheses for a logical rule task were either tautologies (true for all individual cases) or contradictions (true for no individual statements).

Here we confront the assumption that a compositional prior over complex programs must necessarily take the form of a (probabilistic) grammar over a symbolic (probabilistic) language of thought \citep{Piantadosi2016a,Frankland2020}. While a formal language theorist would recognize strings of symbols, particularly bound and unbound variables, as one way among many to encode a universal programming language, symbol strings are the modality in which most programmers express most programs.  The cognitive science community has thus often assumed that \emph{if} the brain has simulation and reasoning capabilities equivalent to a universal computer, \emph{then} it must employ a serialized, symbolic representation \cite{marcus2018algebraic}.  We provide a counterexample in which compositionality is expressed via network structure.
\vspace{-1em}
\paragraph{Contributions} Here we contribute an alternative generative model for representing programs: free categories.  From this structure, we can sample ``correct by construction'' probabilistic programs, whose support must contain the data, without maintaining a symbol table or performing additional type-checking steps.  We show how our formalism allows neural networks to serve as primitives in probabilistic programs, similarly to \cite{Valkov2018}.  We learn both program structure and model parameters end-to-end.
\vspace{-0.5em}
\section{The free category prior over programs}
\begin{figure}[t!]
    \centering
    \includegraphics[width=\columnwidth]{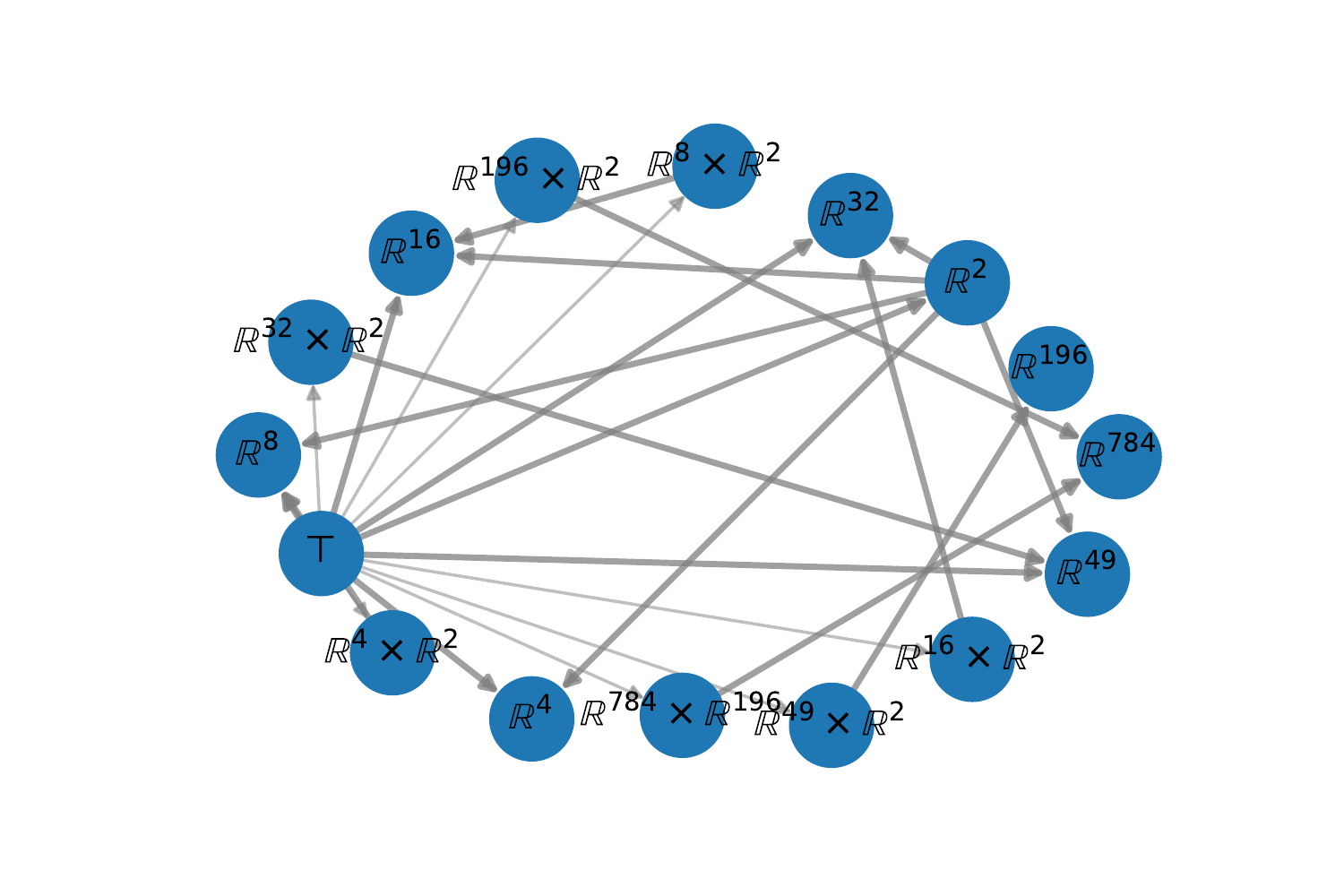}
    \caption{The multigraph (or nerve) of a free category generated by deep generative models.}
    \label{fig:vae_vlae_free_category}
    \vspace{-1.5em}
\end{figure}
\begin{listing*}
    \begin{minted}[escapeinside=||,mathescape=true]{python}
    def path_between(A, B, g, P):
        loc = A
        f = |$id_A$|
        while loc != B:
            probs = [|$P_{a, B}$| for a in g.out_arrows(loc)]
            arrow = sample(Categorical(probs))
            f = f >> arrow # Composition of morphisms
            loc = dest(arrow)
        return f
    \end{minted}
    \caption{The loop for sampling short paths between a source $A$ and a destination $B$ in a free category described by the graph $g$ and biased random walk $P$.}
    \label{alg:path_between}
\end{listing*}
\begin{figure}[t!]
    \centering
    \includegraphics[width=\columnwidth]{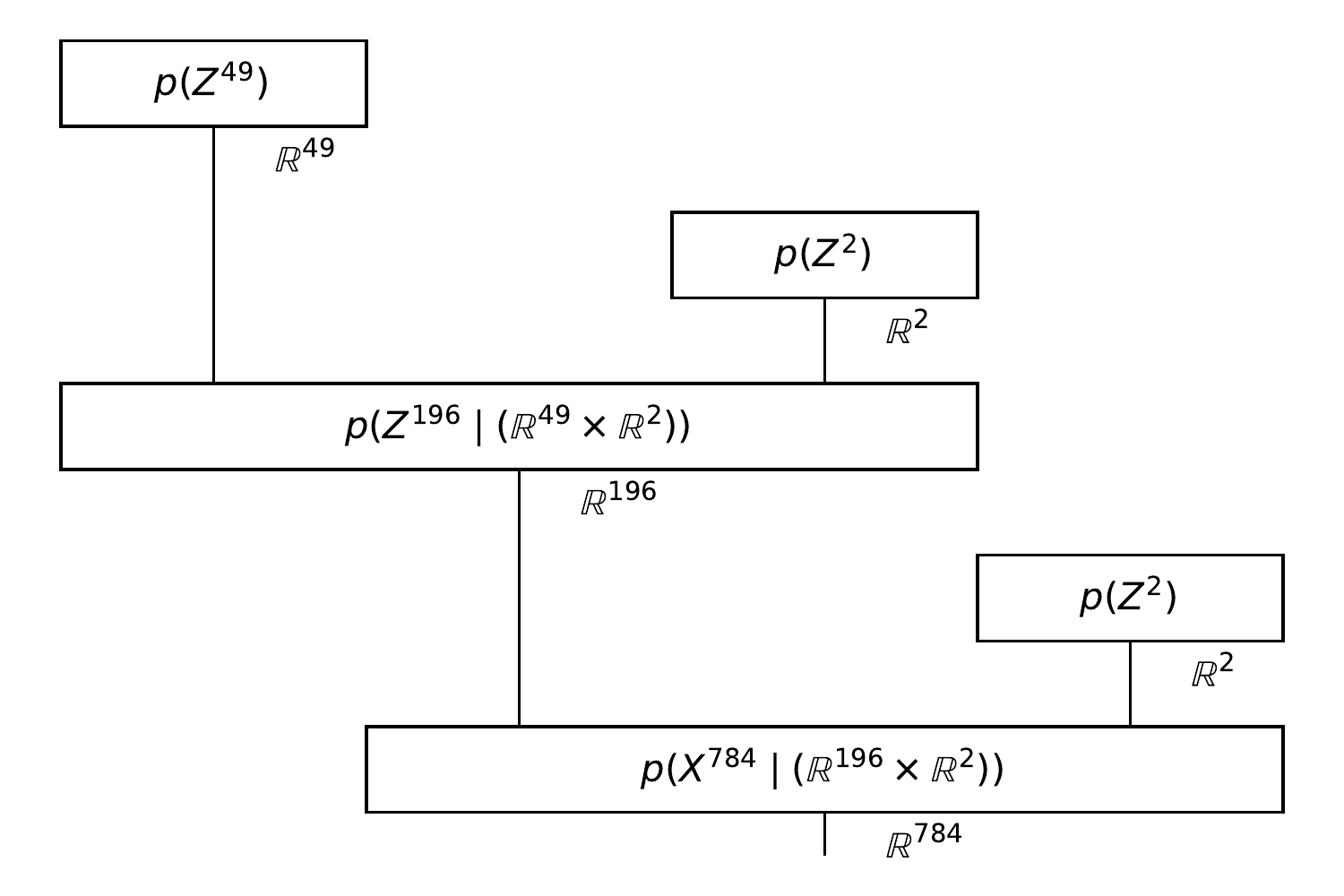}
    \caption{String diagram drawn from the posterior distribution over deep generative model architectures for Omniglot.}
    \label{fig:omniglot_example_morphism}
    \vspace{-1em}
\end{figure}
A category $\mathcal{C}$ consists of a set of \emph{objects} $Ob(\mathcal{C})$, and for every pair of objects $A, B \in Ob(\mathcal{C})$ a set of \emph{morphisms} $\mathcal{C}(A, B)$.  Individual morphisms can be denoted $f: A \rightarrow B$, and each object $A$ is equipped with a unique $id_A: A \rightarrow A$.  As implied by the notation, in programming terms we can imagine objects to be data-types and morphisms to be functions between them.  There are also two major laws which all morphisms must obey to form a category: the associativity of morphism composition, and the identity of unital composition with identity morphisms.

The closure of a directed multigraph under associative composition and self-looping forms the \emph{free} category over that multigraph.  When the number of generating morphisms (edges in the multigraph) is finite, we say that the category is a finitely-generated free category, and we can represent it computationally via its underlying multigraph, $G = (V, E)$.

In Figure~\ref{fig:vae_vlae_free_category}, we show the multigraph of a free, finitely-generated category.  The vertices $V$ correspond to (tuples of) tensor shapes $Ob(\mathcal{C})$, and the edges $E$ correspond to probabilistic decoder networks that act as generating morphisms.  There is a privileged object $\top$, known as the ``terminal'' or ``unit'' object; generating morphisms such as $f: \top \rightarrow \mathbb{R}^{64}$ represent prior distributions $p_f(Z), Z \in \mathbb{R}^{64}$ over the individual vector spaces.  By designing a probabilistic generative model to sample paths through this graph, we thus obtain a prior over variational autoencoder architectures ~\citep{kingma2013auto,Zhao2017}, or, equivalently, a domain-specific language for deep generative models.

Syntactically, we have replaced programs sampled from a grammar with string diagrams sampled from a free category.  These are conventionally read from the top (leaves of an abstract-syntax tree) to the bottom (root of the tree).  An example can be seen in Figure~\ref{fig:omniglot_example_morphism}.

To sample morphisms from the free category, we conduct a biased random walk from a ``source'' object $i$ to a ``destination'' object $j$.  We begin by transforming the directed multigraph $G$ into a directed graph $G'_{\mathcal{C}}=(V', E')$, in which each generating morphism $f: A \rightarrow B$ becomes a vertex $v\in V'$ with a single incoming edge (from $A\in V'$) and a single outgoing edge (to $B\in V'$).  Denoting the adjacency matrix of $G'$ as $A \in \mathbb{R}^{|V'| \times |V'|}$, we can now construct transition probabilities $P_{i,j}$ for all $(i, j) \in |V'| \times |V'|$,
\begin{align}
    \log P_{i,j} &\propto \frac{1}{\beta} \left(e^{A} + AW\right)_{i,j}, \\
    p(i\rightarrow j \mid \beta, W) &= \text{softmax}\left(\frac{1}{\beta} \left(e^{A} + AW\right)\right)_{i,j}, \\
    &= \frac{\exp \left\{\frac{1}{\beta} \left(e^{A} + AW\right)_{i,j} \right\} }{\sum_{j=1}^{|V'|}\exp \left\{\frac{1}{\beta} \left(e^{A} + AW\right)_{i,j}\right\}}.
\end{align}
Here, we write $e^{A}$ for the matrix exponential of $A$, which describes the ``long-run'' transition structure of an unbiased random walk, and we also incorporate weights $W\in \mathbb{R}^{|V'| \times |V'|}$ that can encode arbitrary ``preferences'' for some primitive arrows over others.  The first term inside the parentheses thus imposes an inductive bias towards primitive arrows that belong to many multi-step paths through the graph.  The second term incorporates a learnable preference for some edges over others.  The inverse temperature $\beta > 0$ denotes the relative ``confidence'' of the soft optimization.
Rather than using $-\log P_{i,j}$ as an ``intuitive distance''~\citep{baram2018intuitive,Behrens2018} that compresses shared path segments, we can also hold $j$ constant across a whole path and use $-\log P_{\cdot,j}$ as an intuitive distance to the destination node.  This observation inspires the core of our generative model, seen in Listing~\ref{alg:path_between}, and can be considered a stochastic generalization of path-based planning techniques \citep{blum1997fast,Blum2000}.

Since our underlying free category includes Cartesian product objects such as $(A,B)$, we include ``macro'' arrows $\top \rightarrow (A, B)$ in the multigraph over which we perform the random walk.  When sampled, we ``expand'' these macros by sampling morphisms $\top \rightarrow A$ and $\top \rightarrow B$ and applying the Cartesian product $\times$ on morphisms.  The same type of macro can be used to represent exponential objects $B^A$ (the internal hom-set of morphisms $A \rightarrow B$), by recursively invoking the probabilistic subroutine \texttt{path\_between()}.
\vspace{-1em}
\paragraph{The complete generative model} For data $X\in\mathcal{X}$ and latent variables $Z$ sampled by the parameterized morphism $f_\theta$, itself drawn from the free category of the directed multigraph $G$, the complete generative model is
\begin{multline}
    p_\theta(X, Z, f, \beta, W) = p_{\theta}(X \mid Z, f) p_{\theta}(Z \mid f) \\ p_{G}(f \mid \beta, W) p(\beta) p(W),
\end{multline}
where the prior over the inverse temperature $\beta$ and edge weights $W$ are Gamma distributions,
\begin{align*}
    \beta &\sim \gamma (1, 1), \\
    W &\sim \gamma (1, 1).
\end{align*}
\paragraph{Amortized variational inference in the model}
We implement a data-driven proposal with neural networks,
\begin{align*}
    q_\phi(f, \beta, W \mid X) &= p_{G}(f \mid \beta, W) q_\phi(\beta \mid X) q_\phi(W \mid X).
\end{align*}
We also assume that each primitive morphism $f$ comes equipped with a trainable stochastic inverse~\citep{Stuhlmuller2013} $f^\dagger_\phi: X \rightarrow Z$ (with density $q_\phi(Z \mid X; f)$), so we can derive a complete joint proposal density for all latent variables,
\begin{multline}
    q_\phi(Z, f, \beta, W \mid X) = q_\phi(Z \mid X; f) q_\phi(f, \beta, W \mid X).
\end{multline}
This proposal can then be made to approximate the true Bayesian posterior by optimizing the Evidence Lower Bound (ELBO),
\begin{align}
    \mathcal{L}(\theta, \phi) &= \mathbb{E}_{q}\left[ \frac{p_\theta(X, Z, f, \beta, W)}{q_\phi(Z, f, \beta, W \mid X)} \right].
\end{align}
\vspace{-2em}
\paragraph{Implementation and optimization}
We have implemented this generative model in Pyro~\citep{bingham2019pyro} atop the DisCoPy~\citep{DeFelice2020} library for applied category theory in Python and the NetworkX~\citep{SciPyProceedings_11} library for graph analysis.  NVIL~\citep{mnih2014neural} and graph-based~\citep{schulman2015gradient} gradient estimation techniques were used to perform Stochastic Variational Inference~\citep{hoffman2013stochastic}.

The library implementing the fundamental operations described here is called \href{https://github.com/neu-pml/discopyro}{Discopyro} and is available online, along with the \href{https://github.com/esennesh/categorical_bpl}{code for reproducing our experiments}.  Please note that we regard the exponentiated adjacency matrix $e^{A}$ as a constant for optimization purposes, calculating it once upon construction of the multigraph without allowing gradients to flow through it.  When we have attempted to use differentiable approximate implementations of the matrix exponential, we have found that the approximate posterior distributions collapse down to a single preferred morphism per minibatch of input data due to the extreme effects of small changes to the real-valued adjacency matrix.
\vspace{-0.5em}
\section{Application: variational autoencoder architecture search}
As an example domain, we consider structured variational autoencoder architectures for $28 \times 28$ grayscale image data.  The structures we consider include standard variational autoencoders~\citep{Kingma2013} with Continuous Bernoulli likelihoods~\citep{loaiza2019continuous}, variational ladder autoencoders~\citep{Zhao2017} with continuous Bernoulli likelihoods, and the spatially-invariant Attend-Infer-Repeat variant \citep{crawford2019spatially} with Gaussian likelihoods.  All proposals and priors over vector latent variables are Gaussian.

We list the combinations of network architecture and dataset tested in Table~\ref{tab:datasets_and_architectures}.  We include MNIST~\citep{lecun2010mnist} for a standard of comparison, Fashion MNIST~\citep{xiao2017/online} for a more up-to-date benchmark, and Omniglot~\citep{Lake2015} as a challenge dataset for learning from few examples per class.  The former two datasets were split 90/10 into training and validation data, while Omniglot provided a full test-set of handwritten characters not present in the training dataset.  All evaluations are performed on the validation and test sets, respectively.  Minibatching was used to subsample independent, identically distributed samples from each class within each dataset, treating alphabets as classes in Omniglot.

Learning a latent-variable model over $28\times 28$ images amounts to sampling a morphism $f: \top \rightarrow \mathbb{R}^{784}$ whose path-length in the underlying multigraph is at least two (which we enforce via extra looping and edge-filtering conditions on the code in Listing~\ref{alg:path_between}).
\begin{table}
    \centering
    \begin{tabular}{c|c|c|c}
        \toprule
              & VAE & VLAE & SPAIR \\
        \midrule
        MNIST & \cmark & &  \\
        \midrule
        Fashion MNIST & & \cmark & \\
        \midrule
        Omniglot & & \cmark & \cmark \\
        \bottomrule
    \end{tabular}
    \caption{Dataset-architecture combinations}
    \label{tab:datasets_and_architectures}
\end{table}
\vspace{-0.5em}
\subsection{Results} 
\begin{figure}[t!]
    \includegraphics[width=\columnwidth]{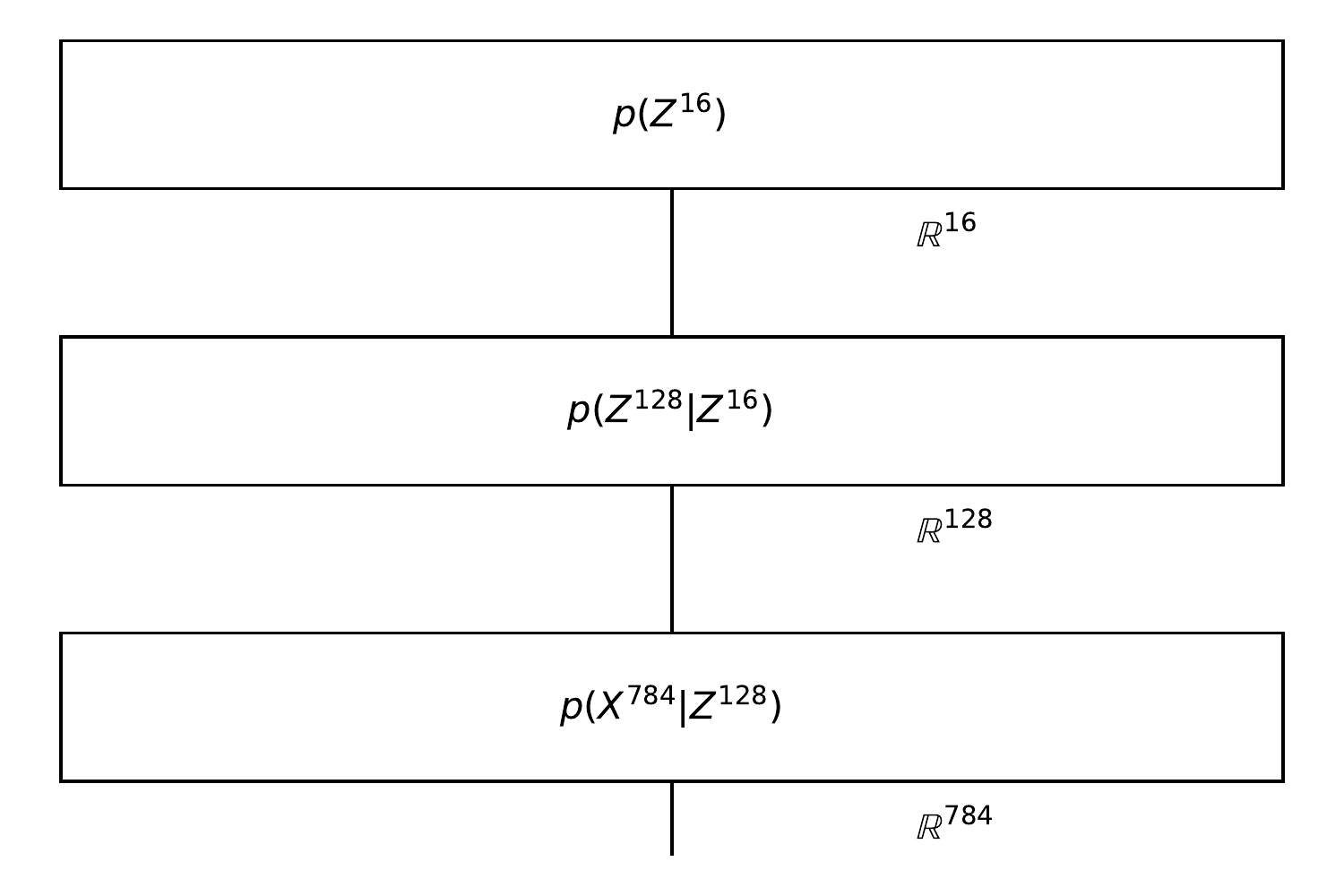}
    \caption{String diagram with vanilla VAE architecture sampled from MNIST's approximate posterior.}
    \label{fig:mnist_string_diagram}
    \vspace{-1em}
\end{figure}

\begin{figure}[t!]
    \includegraphics[width=\columnwidth]{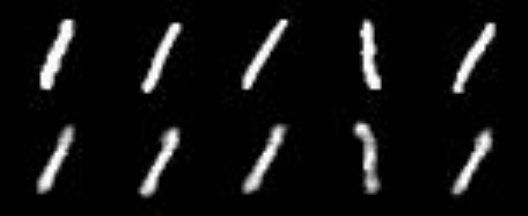}
    \caption{\textbf{Above}: original MNIST digits. \textbf{Below}: Reconstructions from the learned model.}
    \label{fig:mnist_reconstructions}
    \vspace{-1em}
\end{figure}

\begin{figure}[t!]
    \includegraphics[width=\columnwidth]{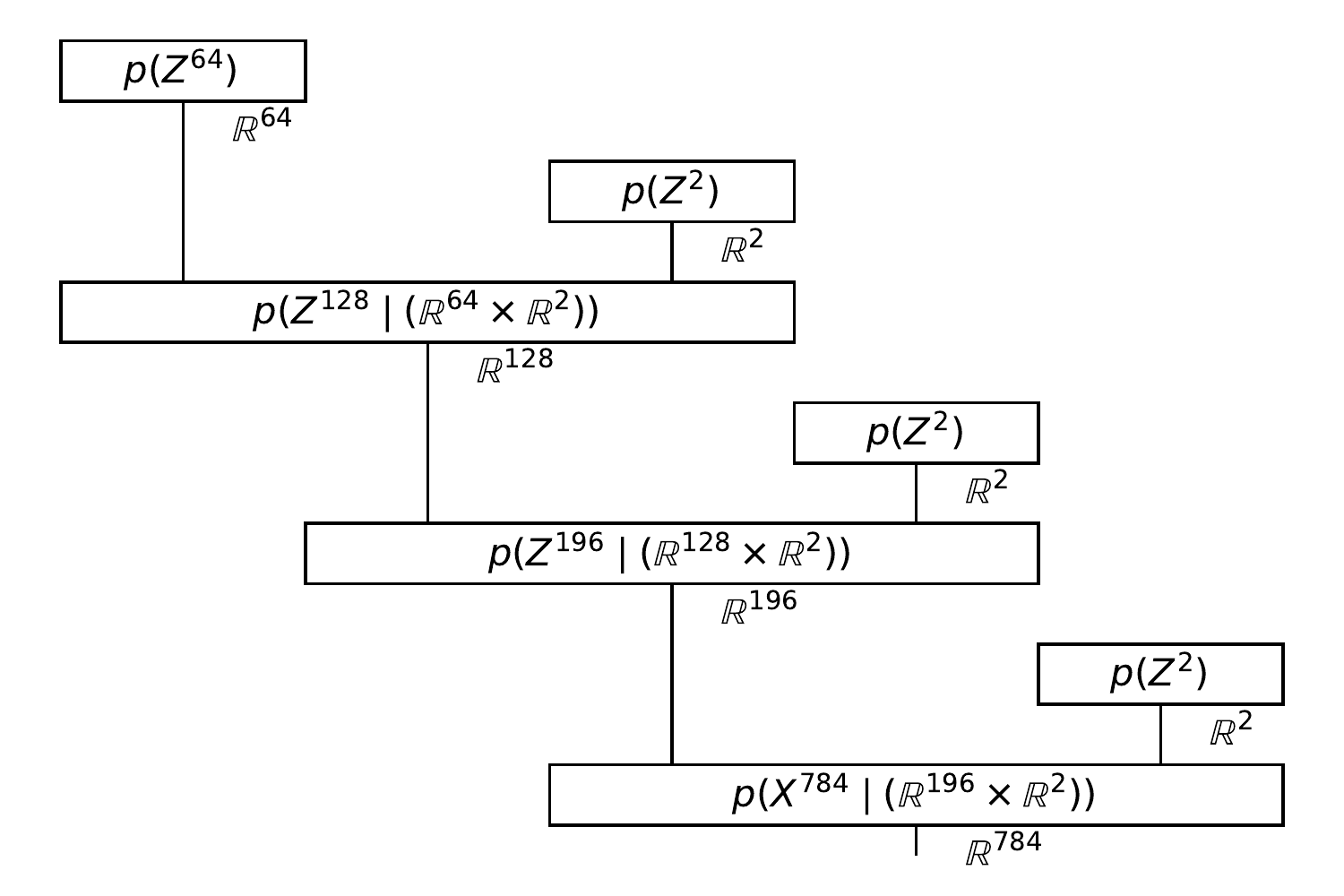}
    \caption{String diagram with VLAE architecture sampled from Fashion MNIST's approximate posterior.}
    \label{fig:fashion_mnist_string_diagram}
    \vspace{-1.5em}
\end{figure}

\begin{figure}[t!]
    \includegraphics[width=\columnwidth]{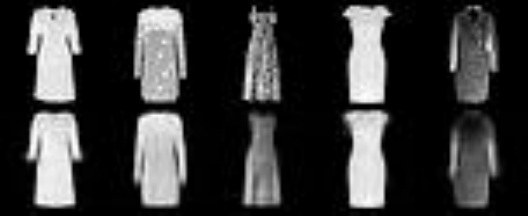}
    \caption{\textbf{Above}: original Fashion MNIST images. \textbf{Below}: Reconstructions from the learned model.}
    \label{fig:fashion_mnist_reconstructions}
    \vspace{-1.5em}
\end{figure}

\begin{figure*}[t!]
    \centering
    \includegraphics[width=\columnwidth]{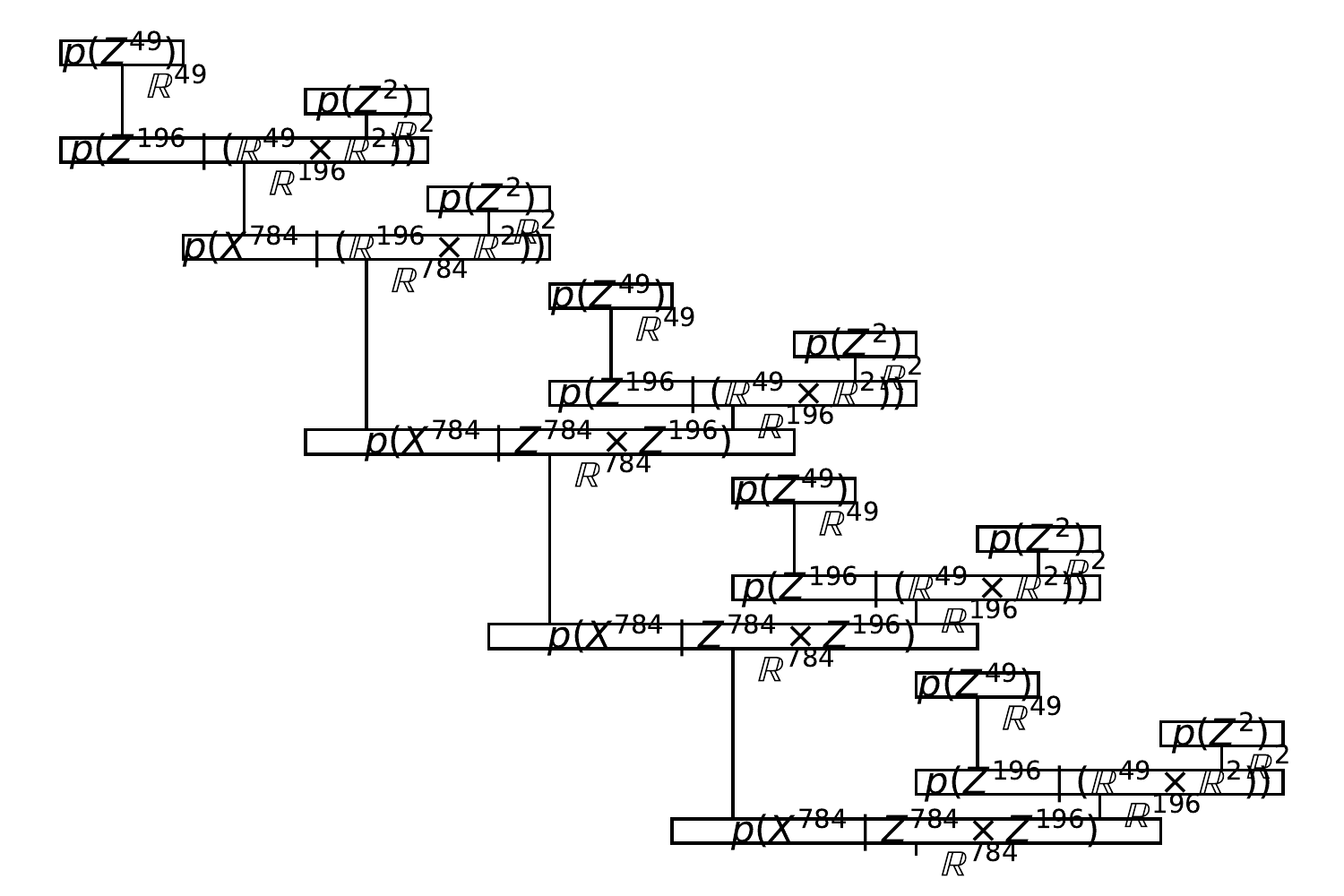}
    \caption{String diagram sampled from Omniglot's approximate posterior.}
    \label{fig:omniglot_spair_string_diagram}
    \vspace{-1em}
\end{figure*}

\begin{figure}[t!]
    \includegraphics[width=\columnwidth]{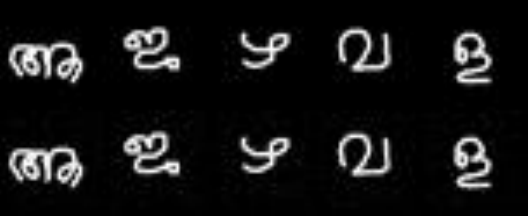}
    \caption{\textbf{Above}: original Omniglot characters. \textbf{Below}: Reconstructions from the learned SPAIR model.}
    \label{fig:omniglot_spair_reconstructions}
    \vspace{-1em}
\end{figure}

\begin{figure*}[t!]
    \centering
    \includegraphics[width=\columnwidth]{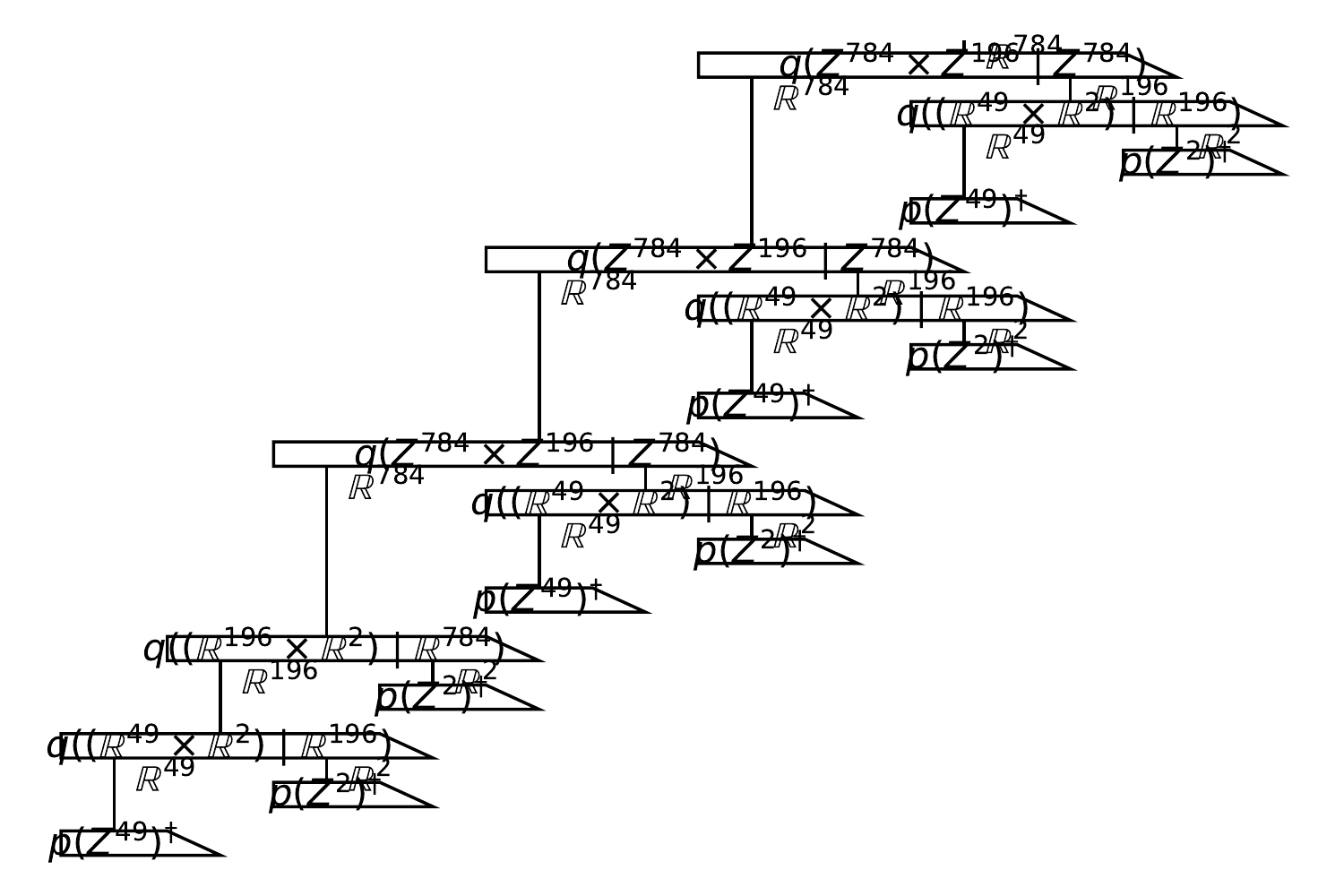}
    \caption{Dagger diagram learned as stochastic inverse of Figure~\ref{fig:omniglot_spair_reconstructions}.}
    \label{fig:omniglot_spair_dagger_diagram}
    \vspace{-1em}
\end{figure*}

Table~\ref{tab:elbo_results} shows training times and validation losses.
\begin{table}
    \centering
    \begin{tabular}{c|c|c|c}
        \toprule
              & Epochs & $\mathcal{L}$ (validation) & $\mathcal{L}$ (test) \\
        \midrule
        MNIST & 500 & $8.36\times 10^5$ & n/a \\
        \midrule
        Fashion MNIST & 500 & $4.87 \times 10^5$ & n/a \\
        \midrule
        Omniglot & 1200 & n/a & $1.46 \times 10^5$ \\
        \bottomrule
    \end{tabular}
    \caption{Evidence lower bounds on validation and test data for our three example datasets}
    \label{tab:elbo_results}
\end{table}
\vspace{-1em}
\paragraph{MNIST} Figure~\ref{fig:mnist_string_diagram} shows a VAE architecture sampled from the amortized approximate posterior over the MNIST validation data. Figure~\ref{fig:mnist_reconstructions} compares the original test images and their reconstructions.  Reconstructions are satisfactory.
\vspace{-1em}
\paragraph{Fashion MNIST} Figure~\ref{fig:fashion_mnist_string_diagram} shows a learned VLAE architecture sampled from the amortized approximate posterior over the Fashion MNIST validation data.  Figure~\ref{fig:fashion_mnist_reconstructions} compares the original test images and their reconstructions.  Reconstructions capture only high-level shape and texture.
\vspace{-1em}
\paragraph{Omniglot} Figure~\ref{fig:omniglot_spair_string_diagram} shows a learned architecture sampled from the amortized approximate posterior over the Omniglot evaluation data.  The architecture makes use of a spatial attention mechanism \citep{Jaderberg2015} to improve learning and reconstruction, and comes from the same approximate posterior distribution as the VLAE architecture above.  Figure~\ref{fig:omniglot_spair_dagger_diagram} diagrams the associated dagger morphism. Figure~\ref{fig:omniglot_spair_reconstructions} compares the original test images and their reconstructions.  Reconstructions appear to capture the relevant details of each character with high quality, most likely thanks to spatial attention being used to encode glimpses of whole character components in the inference model.
\vspace{-1em}
\section{Discussion}
\paragraph{Future work} We hope that future work can extend what we have presented here in several ways.  The most clearly necessary extension, if the intended application is amortized/autoencoding inference, is to cover nonstochastic morphisms, which do not sample their outputs as random variables.  This would amount to the separation of a type system (in which morphisms map from a source to a destination object) from an effect system (in which morphisms add random variables to a trace), with a more complex form of the stochastic planning-graph techniques presented here.

An applied extension of this work would consider would consider an challenging concept-learning benchmark designed, such as ARC~\citep{Chollet2019}, CURI~\citep{vedantam2020curi}, or Bongard-LOGO~\citep{nie2020bongard}.  Any program-learning problem with a well-defined, differentiable likelihood could be considered within our framework, including those in which intermediate latent variables are computed deterministically rather than sampled.

This work could also be extended by passing from a strictly finite multigraph underlying the free category prior to an infinite multigraph.  This infinite multigraph would include structure created by (recursively) applying endofunctors (such as universal constructions or polymorphic algebraic data types) and natural transformations to the base category, and stochastic path-planning would take account of this structure.  This would extend the free category prior to a Bayesian nonparametric model, and end-to-end training would then require truncated or Russian Roulette~\citep{xu2019variational} ELBO estimation.
\vspace{-1em}
\paragraph{Conclusions} We have presented the free category prior, a neuro-categorical model of compositional program learning in custom domain-specific languages.  Our generative model allows for sampling programs which are well-typed, and therefore correct-by-construction in the sense of assigning a nonzero, non-unity likelihood to the data.  We emphasize that our generative model does not involve any rejection-sampling steps, nor other complex deterministic heuristics.

We have applied the free category prior to learning modular variational autoencoder architectures for three common datasets, and have shown that the primitives in a free category's domain-specific language can be parameterized and learned end-to-end by variational Bayes.
        
\bibliographystyle{aaai}
\bibliography{aaai20}

\end{document}